%% file: main_arxiv.tex
\documentclass{article} 
\usepackage{iclr2026_conference,times}
\usepackage{graphicx}
\input{math_commands.tex}

\input{env.tex}

\usepackage{hyperref}
\usepackage{url}
\usepackage{booktabs}
\usepackage{multirow}

\usepackage{xcolor}
\definecolor{GREY}{rgb}{0.6, 0.6, 0.6}
\definecolor{YELLOW}{rgb}{1,0.6,0} 

\usepackage{tcolorbox}
\usepackage{enumitem}
\usepackage{wrapfig}

\tcbuselibrary{skins}

\newcommand{\method}{\textsc{Deg-Rag}}

\title{Less is More: Denoising Knowledge Graphs For Retrieval Augmented Generation}

\author{
Yilun Zheng$^{1}$\thanks{Equal contribution.}, 
Dan Yang$^{1}$\footnotemark[1],  
Jie Li$^{1}$,  
Lin Shang$^{2}$,  
Lihui Chen$^{1}$,  
Jiahao Xu$^{1}$,  
Sitao Luan$^{3,4}$\thanks{Corresponding author.} \\
$^1$Nanyang Technological University, \quad
$^2$Nanjing University, \quad
$^3$Mila, Quebec AI Institute, \\
$^4$University of Montreal\\
\texttt{Email: yilun001@e.ntu.edu.sg, luansito@mila.quebec}
}

\iclrfinalcopy 
\begin{document}

\maketitle

\begin{abstract}


Retrieval-Augmented Generation (RAG) systems enable large language models (LLMs) instant access to relevant information for the generative process, demonstrating their superior performance in addressing common LLM challenges such as hallucination, factual inaccuracy, and the knowledge cutoff. Graph-based RAG further extends this paradigm by incorporating knowledge graphs (KGs) to leverage rich, structured connections for more precise and inferential responses. A critical challenge, however, is that most Graph-based RAG systems rely on LLMs for automated KG construction, often yielding noisy KGs with redundant entities and unreliable relationships. The redundancy not only degrades retrieval and generation performance but also increases computational cost. Crucially, current research does not comprehensively address the denoising problem for LLM-generated KGs. In this paper, we introduce DEnoised knowledge Graphs for Retrieval Augmented Generation (\method), a framework that addresses these challenges through: (1) entity resolution, which eliminates redundant entities, and (2) triple reflection, which removes erroneous relations. Together, these techniques yield more compact, higher-quality KGs that significantly outperform their unprocessed counterparts. Beyond these methods, we conduct a systematic evaluation of entity resolution for LLM-generated KGs, examining different blocking strategies, embedding choices, similarity metrics, and entity merging techniques. To the best of our knowledge, this is the first comprehensive exploration of entity resolution in LLM-generated KGs. Our experiments demonstrate that this straightforward approach not only drastically reduces graph size but also consistently improves question answering performance across diverse popular Graph-based RAG variants. Code is available at \href{https://github.com/157114/Denoise}{(https://github.com/157114/Denoise)}.

\end{abstract}

\input{sections/Introduction}

\input{sections/related_work}

\input{sections/Preliminary}

\input{sections/Denoising_KG}

\input{sections/Experiments}

\vspace{-0.2cm}
\section{Conclusion and Future Works}
\vspace{-0.2cm}

In this work, we investigated how denoising LLM-generated KGs benefits Graph-based RAG. We introduced \method, which combines entity resolution and triple reflection to remove redundant entities and filter unreliable relations. Across four Graph-based RAG variants and four datasets, \method~reduces around half the size of the entities and relations while preserving or improving QA quality and lowering storage cost. Our component analysis shows that type-aware blocking is consistently strong, classical KG embeddings such as ComplEx can rival LLM embeddings, ego information is essential and neighbor cues help in some settings, and direct merging generally outperforms synonym-only linking. Hyperparameter sweeps reveal wide operating regimes and sometimes allow up to 70\% entity reduction without hurting performance. Our methods focus on improving the quality of KGs and can be used alongside advances in knowledge-graph-based LLM applications \citep{KG_LLM_app1,KG_LLM_app2,KG_LLM_app3}.

While effective, \method~has limitations. Our study uses four QA datasets and non-large-scale KGs. Triple reflection depends on LLM prompting and the LLM-as-judge setup, which can introduce calibration bias. Gains are bounded by attribute richness. For example, graphs with only short names without rich descriptions limit resolution quality. In future work, we will extend \method~to more datasets and larger-scale KGs, generalize the denoising pipeline to other LLM-generated data structures beyond KGs, and richer evaluations beyond LLM as judges.


\section*{Acknowledgment}
Supported by the Overseas Open Funds of State Key Laboratory for Novel  Software Technology of China ( No.KFKT2025A06)
\section*{Reproducibility statement}
We have provided the codebase in supplementary material and all the results in this paper are reproducible. The additional implementation details and experimental setups can be found in Section \ref{sec:setup} and Appendix~\ref{apd:sec_experimental_details}.

\section*{Ethics statement}
All of the authors in this paper have read and followed the ethics code.

\bibliography{iclr2026_conference}
\bibliographystyle{iclr2026_conference}

\clearpage
\appendix
\section*{The Use of Large Language Models}
In this work, we employed LLMs as auxiliary tools to support the preparation of the manuscript. Specifically, LLMs were used in two ways: (i) to polish the writing style of the paper by refining grammar, clarity, and readability without altering the technical content, and (ii) to assist in identifying relevant related work by suggesting potential references. Note that LLMs were not involved in designing experiments, analyzing results, or drawing conclusions; these aspects of the study were carried out independently by the authors.

\input{appendix/data_statistics}

\input{appendix/more_experiments}

\input{appendix/experimental_details}

\input{appendix/proof}

\end{document}

%% file: math_commands.tex
\usepackage{amsmath,amsfonts,bm}

\def\eqref#1{equation~\ref{#1}}

\def\1{\bm{1}}

\DeclareMathAlphabet{\mathsfit}{\encodingdefault}{\sfdefault}{m}{sl}
\SetMathAlphabet{\mathsfit}{bold}{\encodingdefault}{\sfdefault}{bx}{n}



%% file: env.tex
\newcommand\ie{\textit{i.e.,}}
\newcommand\eg{\textit{e.g.,}}

\newcommand{\beq}{\begin{equation}}
\newcommand{\eeq}{\end{equation}}
\newcommand{\beqnn}{\begin{equation*}}
\newcommand{\eeqnn}{\end{equation*}}
\newcommand{\beqy}{\begin{eqnarray}}
\newcommand{\eeqy}{\end{eqnarray}}
\newcommand{\beqynn}{\begin{eqnarray*}}
\newcommand{\eeqynn}{\end{eqnarray*}}
\newcommand{\bit}{\begin{itemize}}
\newcommand{\eit}{\end{itemize}}
\newcommand{\ben}{\begin{enumerate}}
\newcommand{\een}{\end{enumerate}}
\newcommand{\bex}{\begin{example}}
\newcommand{\eex}{\end{example}}

\newcommand{\balg}[1]{\begin{algorithm} \caption{#1}}
\newcommand{\ealg}{\end{algorithm}}

\newcommand{\balgc}{\begin{algorithmic}[1]}
\newcommand{\ealgc}{\end{algorithmic}}

\newcommand{\bary}{\begin{array}}
\newcommand{\eary}{\end{array}}
\newcommand{\bmx}{\begin{bmatrix}}
\newcommand{\emx}{\end{bmatrix}}
\newcommand{\bsmx}{\left[\begin{smallmatrix}}
\newcommand{\esmx}{\end{smallmatrix}\right]}
\newcommand{\bmxc}[1]{\left[\begin{array}{@{}#1@{}}}
\newcommand{\emxc}{\end{array}\right]}
\newcommand{\bcn}{\begin{center}}
\newcommand{\ecn}{\end{center}}

\newenvironment{theorem}[2][Proposition]{\begin{trivlist}
		\item[\hskip \labelsep {\bfseries #1}\hskip \labelsep {\bfseries #2.}]}{\end{trivlist}}

%% file: sections/Introduction.tex
\vspace{-0.3cm}
\section{Introduction}\vspace{-0.3cm}
\label{sec:introduction}
Large Language Models (LLMs) have made significant progress in natural language processing, understanding and reasoning \citep{llm_survey,jin2025rl}. However, their capabilities are limited by the delayed access to up-to-date information, susceptibility to hallucination, and weak long-term memory \citep{llm_survey,llm_hallucination,llm_memory}. To mitigate these issues, Retrieval-Augmented Generation (RAG) \citep{rag_original} has emerged to ground LLMs with external knowledge. Given a user query, a RAG system retrieves relevant information from a knowledge base, augments the query with the retrieved context, and then generates a response. RAG enables LLMs to access updated facts, and rapidly adapt to new domain knowledge.
\begin{figure}[htbp]
    \centering
    \includegraphics[width=0.8\linewidth]{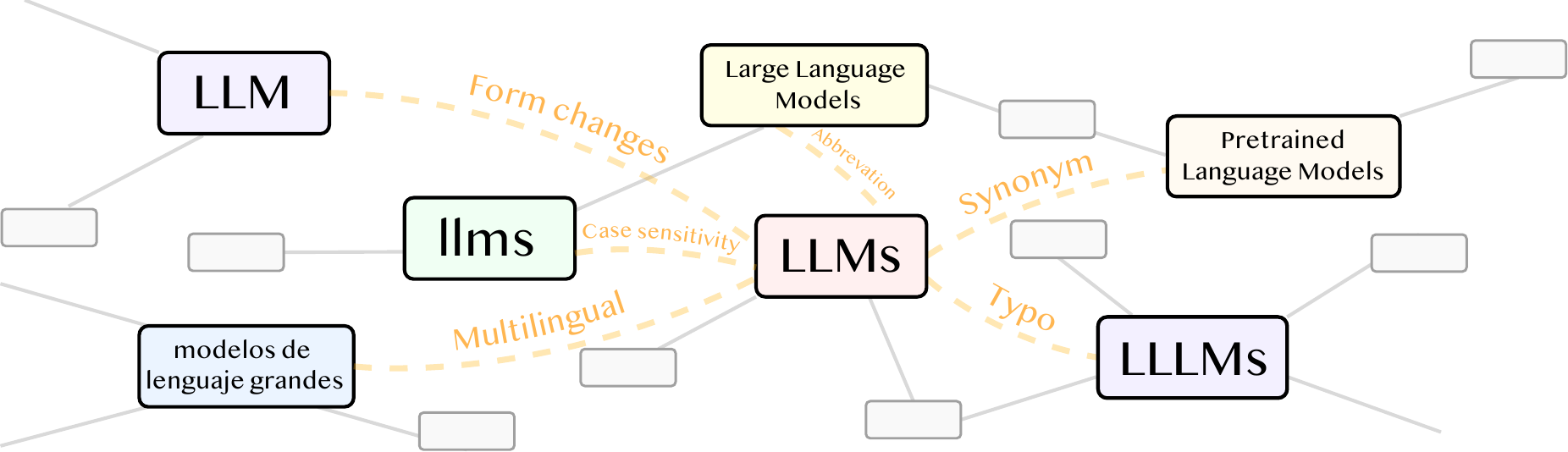}
    \caption{Redundant concept synonyms for ``LLMs'' in a knowledge graph. \textcolor{YELLOW}{Orange dashed lines} indicate synonymic equivalences showing why these entities convey the same meaning as ``LLMs''.
    }
    \label{fig:synonym_example}
\end{figure}

Traditional RAG systems \citep{traditional_rag} retrieve isolated text chunks and ignore relationships among them, which weakens multi-hop reasoning \citep{multihop_reasoning} and overall coherence \citep{rag_cohesiveness}. Graph-based RAG \citep{ms_graphrag, lightrag, hipporag} addresses it by structuring knowledge as a graph and retrieving over that structure. Connectivity among entities allows models to consider inter-document relations rather than treating units as independent chunks, enabling fine-grained, relation-aware retrieval \citep{external_knowledge_relations}. 

As we all know, the quality of graph is critical to the success of graph mining \citep{graph_quality_fundamentals, luan2024heterophilic, zheng2025let}, and many graph-based RAG systems focus on constructing knowledge graphs (KGs) from corpora with LLMs. However, the resulting graphs are often noisy and redundant \citep{llm_kg_quality}. During entity and relation extraction, unlike human experts who can accurately recall and connect new concepts to previously identified entities, LLMs often struggle to consistently maintain earlier entities and relations due to limited long-context capabilities, which leads to duplicates \citep{llm_NER_failure}. As illustrated in Figure \ref{fig:synonym_example}, the extracted entity ``LLMs'' may co-occur with its variants that represent the same concept, \eg{} ``LLM'' (morphology), ``llms'' (casing), ``modelos de lenguaje grandes'' (multilingual), and ``Large Language Models'' (abbreviation expansion). Existing methods, including LightRAG \citep{lightrag}, MS GraphRAG \footnote{To avoid ambiguity, we use MS GraphRAG to refer to the specific GraphRAG method proposed in \citep{ms_graphrag}, and Graph-based RAG to refer to the general class of approaches that leverage knowledge graphs.} \citep{ms_graphrag}, and HippoRAG \citep{hipporag}, typically rely on string-matching heuristics to merge similar entities, leaving many duplicates unresolved. These \textbf{redundant entities} inflate storage, degrade retrieval efficiency and precision. Besides, some outdated and incorrect facts in external corpora\citep{misinformation_llm,llm_error_accumulation, llm_error_accumulation2} yield  \textbf{erroneous triples} in LLM-generated graphs, which mislead retrieval and generation.

To simultaneously reduce the size and improve the quality of generated graphs, we propose DEnoised knowledge Graphs for Retrieval Augmented Generation (\method), which takes entity resolution to remove redundancy, and triple reflection to filter erroneous relations in LLM-generated knowledge graphs for RAG. Entity resolution identifies and links records that refer to the same entity \citep{entity_resolution_deep} and is widely used in traditional KG consolidation \citep{kg_merging}. We conducted comprehensive evaluation and studies tailored to Graph-based RAG, spanning different blocking, entity-embedding, matching, and merging strategies. 
    
Our experiments show that, while removing $40\%$ of the entities and relations in LLM-generated KGs, \method~consistently improves the performance of four representative Graph-based RAG approaches, underscoring the importance of KG quality over its size. We further study the design of different components comprehensively and come up with some interesting findings, \eg{} type-aware blocking is the most effective blocking method, traditional KG embeddings can rival LLM embeddings, neighborhood-based similarity sometimes outperforms ego-based measures, and simple merging often surpasses synonym-edge addition. Together, these findings offer practical guidance for constructing high-quality LLM-generated KGs and for developing more efficient and accurate Graph-based RAG systems, with potential extensions to a wide range of KG-based LLM applications \citep{KG_LLM_app1,KG_LLM_app2,KG_LLM_app3}. In summary, our contributions are as follows:
\begin{itemize}
\item We propose \method, which leverages entity resolution and triple reflection to reduce graph size while improving KG quality for better Graph-based RAG.
\item To the best of our knowledge, we are the first to conduct a comprehensive study of entity resolution for Graph-based RAG, implementing and evaluating different components, including blocking, entity-embedding, matching, and merging strategies.
\item Our experiments demonstrate that \method~improves the performance of four graph-based RAG methods across four benchmark QA datasets by removing approximately $40\%$ of entities and relations. We further analyze how different components of entity resolution contribute to Graph-based RAG performance.
\end{itemize}






%% file: sections/related_work.tex
\vspace{-0.2cm}
\section{Related Work}
\vspace{-0.2cm}
\label{sec:related_work}
Retrieval Augmented Generation (RAG) enables Large Language Models (LLMs) to utilize updated information \citep{llm_update_rag}, access domain-specific knowledge \citep{llm_domain_rag}, and reduce hallucinations \citep{llm_hallucination}. Traditional RAG systems \citep{traditional_rag} organize external knowledge as isolated database chunks, which limits performance in complex reasoning \citep{multihop_reasoning,jiang2024longrag} and contextual completeness \citep{lu2025hichunk,zhong-etal-2025-mix}. To address these limitations, Graph-based RAG presents external information as graphs, retrieving relevant data by considering inter-relationships \citep{graph_rag_survey}. MS GraphRAG \citep{ms_graphrag} constructs communities and generates answers based on community summaries, while LightRAG \citep{lightrag} retrieves relevant entities, relationships, and subgraphs using keywords from queries. HippoRAG \citep{hipporag} employs PageRank \citep{pagerank} for efficient entity retrieval. KAG \citep{KAG} integrates knowledge graphs (KGs) with LLMs through logical-form-guided reasoning, knowledge alignment, and fine-tuning. Despite these advancements, the quality of LLM-generated KGs remains a challenge, as they are often redundant and noisy, hindering efficient knowledge storage and high-quality generation \citep{DIGIMON}.

Entity resolution, which links data records referring to the same real-world entity, is crucial for constructing high-quality KGs \citep{pujara2016generic,eager}. Existing approaches fall into three categories: (1) Traditional methods use string similarity \citep{ER_string_survey,ER_string}, heuristic rules \citep{ER_rule,ER_rule2}, or manually designed schemas \citep{ER_schema} to identify equivalent entities. These methods are computationally efficient and interpretable but struggle with noisy, incomplete, or multilingual data. (2) Embedding-based methods represent entities in continuous vector spaces, matching based on representation similarity. This includes LLM-based embeddings \citep{ER_llm} and KG embeddings like TransE \citep{TransE}, DistMult \citep{DistMult}, and ComplEx \citep{ComplEx}, as well as Graph Neural Networks (GNNs)-based approaches \citep{R_GCN}. These techniques capture structural dependencies across graphs, offering robustness over heuristic methods. (3) LLM-based methods leverage LLMs through prompting \citep{ER_llm_prompt} or fine-tuning \citep{ER_finetune} to identify semantically equivalent entities, providing strong generalization capabilities, though they require careful design for scalability and reliability.

Although many entity resolution methods exist, few focus on improving LLM-generated KG quality. For example, MS GraphRAG \citep{ms_graphrag} and LightRAG \citep{lightrag} use simple string matching for duplicate entity identification. HippoRAG \citep{hipporag} introduces synonym relations based on cosine similarity, and KAG \citep{KAG} predicts synonym relations from one-hop neighbors, merging entities accordingly. However, the impact of enhancing KG quality on Graph-based RAG is largely unexplored. This paper systematically investigates how different entity resolution methods affect the performance of Graph-based RAG, alongside triple reflection, contributing uniquely beyond previous studies.

%% file: sections/Preliminary.tex
\vspace{-0.2cm}
\section{Preliminaries}\label{sec:preliminaries}
\vspace{-0.2cm}
In this section, we introduce the notations and the process of Graph-based RAG.
Given a set of external documents $\mathcal{D} = [d_1, d_2, \dots, d_N]$, Graph-based RAG constructs a knowledge graph (KG) $\mathcal{G} = (\mathcal{E}, \mathcal{R}, \mathcal{T}, \mathcal{A})$, where $\mathcal{E}$, $\mathcal{R}$, and $\mathcal{T}$ denote the sets of entities, relation types and triples, and $\mathcal{A}$ represents the textual description for each entity. The neighbors of an entity $e \in \mathcal{E}$ are defined as the set of entities $\mathcal{N}(e)$ that are directly connected to $e$ through relation $r \in \mathcal{R}$:
\begin{equation}
\mathcal{N}(e) = \{\, e' \in \mathcal{E} \;\mid\; (e, r, e') \in \mathcal{T} \;\;\vee \;\; (e', r, e) \in \mathcal{T}, \; r \in \mathcal{R} \,\}.
\end{equation}
Then, given a user query $Q$, the RAG system (1) retrieves relevant contents from $\mathcal{G}$ via a retrieval function $\mathcal{R}(\cdot)$, (2) augments the query $Q$ with retrieved context using an augmentation function $\text{Aug}(\cdot)$, and (3) generates the final answer $\mathcal{Y}$ with LLMs $\mathcal{M}$. Formally:
\begin{equation}
\mathcal{Y} \;=\; \mathcal{M} \circ \text{Aug}\big[ Q, \mathcal{R}(Q,\mathcal{G}) \big].
\end{equation}
Specifically, the raw documents $\mathcal{D}$ are first segmented into text chunks $\mathcal{C} = [c_1, c_2, \dots, c_M]$. For each chunk $c_m \in \mathcal{C}$, a LLM-based named-entity recognition function $\mathcal{M}_\text{NER}(\cdot)$ is applied, leads to a set of raw triples, entities, and relations:
\begin{equation}
       \resizebox{0.94\hsize}{!}{$ \mathcal{T}_m \;=\; \mathcal{M}_{\text{NER}}(c_m), \;
        \mathcal{T} \;=\; \bigcup_{m=1}^M \mathcal{T}_m, \;
        \mathcal{E} = \{ e_1, e_2 \mid (e_1, r, e_2) \in \mathcal{T} \}, \;
        \mathcal{R} = \{ r \mid (e_1, r, e_2) \in \mathcal{T} \}.$}
\end{equation}
where each entity $e \in \mathcal{E}$ carries its local textual context $\mathcal{A}(e)$. Here, the LLM extracted $\mathcal{E}$ may contain duplicates, aliases, or simple variations. To construct a coherent KG, a deduplication function $\phi: \mathcal{E}\mapsto\mathcal{E}^*$ is applied, which maps each raw entity to a unique canonical entity $\phi(e)$. Then we have the revised entity, triple, and relation sets as: 
\begin{equation}
    \resizebox{0.94\hsize}{!}{$
        \mathcal{E}^* = \{\phi(e) \mid e \in \mathcal{E}\}, \;
        \mathcal{T}^* = \{(e_1,r,e_2) \mid (e_1,r,e_2)\in\mathcal{T},  e_1\in\mathcal{E}^*,e_2\in\mathcal{E}^*\}, \;
        \mathcal{R}^* = \{ r \mid (e_1, r, e_2) \in \mathcal{T}^* \}
    $}
\end{equation}
For each canonical entity $e^* \in \mathcal{E}^*$, we aggregate the textual description with a merge operator $\oplus$:
\begin{equation}
\mathcal{A}^*(e^*) = \bigoplus_{\{e_i : \phi(e_i) = e^*\}} \mathcal{A}(e_i)
\end{equation}

The final denoised KG is $\mathcal{G}^* = (\mathcal{E}^*, \mathcal{R}^*, \mathcal{T}^*, \mathcal{A}^*)$, enabling more efficient retrieval. 

%% file: sections/Denoising_KG.tex
\vspace{-0.2cm}
\section{Denoising Knowledge Graphs}\label{sec:denoising_kg}
\vspace{-0.2cm}
In most popular Graph-based RAG systems, such as LightRAG \citep{lightrag} and MS GraphRAG \citep{ms_graphrag}, a simple string matching strategy is used as the deduplication function to denoise KGs. However, in this way, entities with the same semantic meaning but different forms, \eg case sensitivity, abbreviation, synonym, multilingual, and typos, will be missed and isolated from each other. This will lead to a coarse and redundant KG that impedes efficient storage and retrieval in Graph-based RAG systems. To enhance the performance of Graph-based RAG by denoising LLM-generated KGs, we propose to remove redundant entities by entity resolution in Section \ref{sec:entity_resolution} and remove unreasonable edges by triple reflection in Section \ref{sec:triple_reflection}. This framework enhances the quality of the KGs while reducing their size.

\vspace{-0.2cm}
\subsection{Entity Resolution}\label{sec:entity_resolution}
\vspace{-0.2cm}
Entity resolution for KGs involves several key steps\citep{ER_survey}, (1) \textbf{Blocking:} partitions raw entities into blocks to minimize the number of entity pairs that need to be compared. (2) \textbf{Matching and Grouping:} identify entities that represent the same real-world object and then put these matched entities into groups representing a single resolved entity. (3) \textbf{Merging and Linking:} combine the raw entities in each cluster into a canonical representation and update the KG by creating or deleting relations as needed. With the above steps, we introduce how to use entity resolution to improve the quality of LLM-generated KGs as follows.

\begin{figure}[h]
    \centering
    \includegraphics[width=0.95\linewidth]{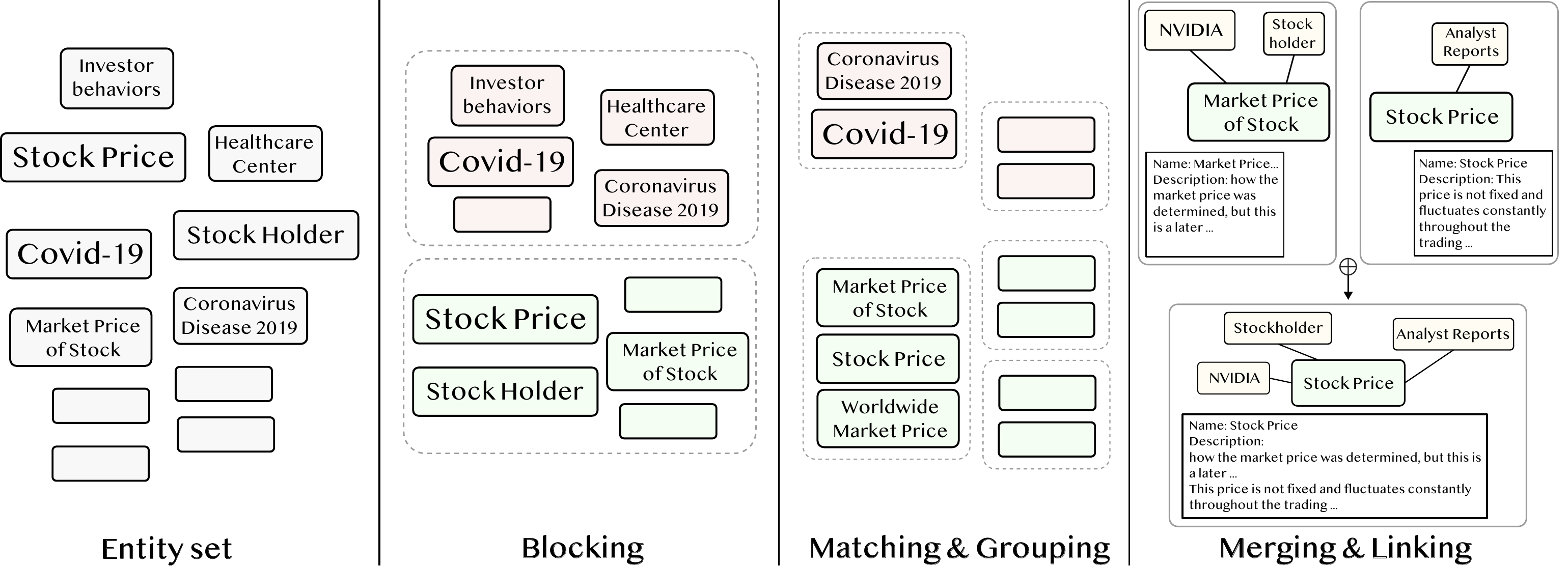}
    \caption{The overall framework of entity resolution for knowledge graphs \citep{ER_survey}.}
    \label{fig:entity_resolution}
\end{figure}


\textbf{Blocking.} To reduce computational costs and unnecessary entity comparisons, blocking is applied to the entity set $\mathcal{E}$ before entity matching \citep{ER_blocking}. Formally, blocking is a mapping
\begin{align}
\text{Block}: \mathcal{E} \;\mapsto\; \mathcal{B} = \{B_1, B_2, \dots, B_K\}, \; \bigcup_{k=1}^K B_k = \mathcal{E}
\end{align}
where each block $B_k$ is a subset of entities that are more likely to be matched. In this paper, we consider three types of blocking strategies: semantic-based, entity type-based, and structural-based \citep{ER_survey}.

(1) Semantic-Based Blocking. Entities are represented as embeddings generated from their descriptions $\mathcal{A}(e)$ using an embedding model $f_{\text{emb}}(\cdot)$. The entity set is partitioned into $k$ clusters by:
\begin{align*}
\mathcal{B} = \texttt{kmeans}\big(\{f_{\text{emb}}(\mathcal{A}(e)) \mid e \in \mathcal{E}\}, k\big), 
\end{align*}
To avoid manual selection of cluster number $k$, we use a rule-of-thumb heuristic $k = \sqrt{\tfrac{|\mathcal{E}|}{10}}$ \citep{kmean_k_selection}. This strategy leverages global semantic similarity but is computationally more expensive for large graphs.

(2) Entity Type-Based Blocking. Entities are first classified into types using a type mapping function $\tau: \mathcal{E} \mapsto \Omega$. Entities with the same type $t \in \Omega$ are grouped into the same block:
\begin{align*}
\mathcal{B} = \{ \{ e \in \mathcal{E} \mid \tau(e) = t \} \;\mid\; t \in \Omega \}.
\end{align*}
If a block contains too many entities, we further subdivide it using $k$-means. The entity type-based blocking limits the matches within the same type of entities, which avoids excessive pair comparisons.

(3) Structural-based Blocking. This strategy exploits graph connectivity under the assumption that semantically similar entities are likely to share neighbors. If an entity $e$ has at least two neighbors, we construct a block for its neighbor set $\mathcal{N}(e)$, and the set of final structural-based blocks is then
\begin{align*}
\mathcal{B} = \{ \mathcal{N}(e) \mid e\in\mathcal{E}, |\mathcal{N}(e)|\geq 2\} 
\end{align*}
This blocking is based on the assumption that entities co-occur as neighbors of the same nodes are more likely to present the same meaning, \eg ``Large Language Models'' and ``Pretrained Language Models'' may be placed in the same block if they both connect to the entity ``GPU'' through the relation ``run on.''. Therefore, the structural context of shared neighbors serves as a strong signal for blocking.



\textbf{Matching and Grouping.}
After blocking, the objective is to identify sets of entities in each block that represent the same concept then group entities with the same meaning. Given a block \( B \subseteq \mathcal{E} \), the matching function derives a partition:
\begin{align}
\text{Match}: B \;\mapsto\; G = \{G_1, G_2, \dots, G_L\}, \; \bigcup_{l=1}^L G_l \subseteq B,
\end{align}
where each $G_l$ is a group of equivalent entities. To match entities, we first obtain the embedding of each entity $h(e)$ in the KG, then select the entity embedding for matching. Specifically, embedding methods used in this paper include KG embeddings: TransE \citep{TransE}, DistMult \citep{DistMult}, and ComplEx \citep{ComplEx}; graph neural network embeddings: CompGCN \citep{CompGCN} and R-GCN \citep{R_GCN}; and LLM embeddings of Qwen3-Embedding-8B \citep{qwen3embedding}. 

To match similar nodes with proper information after embedding, we consider the calculation of the following similarity scores: (1) \textbf{Ego node similarity.} It compares entity embeddings $h(e_i)$ and $h(e_j)$, which is computationally efficient but may miss structural context. (2) \textbf{Neighbor similarity.} It compares averaged neighbor embeddings $\bar{h}_{\mathcal{N}}(e_i)$ and $\bar{h}_{\mathcal{N}}(e_j)$, leveraging structural context to identify entities with similar roles. (3) \textbf{Type-aware Neighbor similarity.} It compares type-specific averaged neighbor embeddings $\bar{h}_{\mathcal{N}_t}(e_i)$ and $\bar{h}_{\mathcal{N}_t}(e_j)$ for each type $t \in \Omega$, where $\mathcal{N}_t(e) = \{e' \in \mathcal{N}(e) \mid \tau(e') = t\}$, then averages across types: $\text{sim}(e_i, e_j) = \frac{1}{|\Omega|} \sum_{t \in \Omega} \text{sim}_t(\bar{h}_{\mathcal{N}_t}(e_i), \bar{h}_{\mathcal{N}_t}(e_j))$. This reduces noises from irrelevant neighbors and enables precise matching within specific entity types, particularly when entities of different types may have fundamentally different embedding distributions. (4) \textbf{Ego+neighbor similarity.} It considers both the ego node and neighbor information by concatenating the embeddings in (1) and (2). (5) \textbf{Ego+Type-aware neighbor similarity.} It considers both the ego node and subset of neighbor information by concatenating the embeddings used in (1) and (3). Each matching method captures different aspects of entity similarity and presents distinct trade-offs.


After matching, entities $e_i$ and $e_j$ are grouped together if their similarity exceeds threshold $\delta_\text{ER}$, and we assign each entity to a group using the function $g: \mathcal{E} \mapsto G$. 

\textbf{Merging or Linking.}  
Once entity groups $G$ are obtained, we finalize the KG $\mathcal{G}^*$ by editing the previous KGs with the following three strategies:

(1) Direct Merging.
This approach first selects a single canonical entity $e_l^*=\phi(G_l)$ given a group $G_l$, where $\phi(\cdot)$ refers to a canonical selection function. In this paper, we use random selection for $\phi(\cdot)$. Then, all the other entities inside the group $G_l$ are merged into the canonical entity $\hat{e}_l$. The KG is updated by appending the descriptions of merged entities to that of the canonical entity, reconnecting their relations to the canonical entity, and removing relations that involve the merged entities. The above process can be expressed as:
\begin{equation} \label{eq:attr_merge}
\mathcal{E}^{*} = \{\phi(G_l) \mid G_l \in G\},  \;
\mathcal{A}^{*}(\phi(G_l)) = \bigcup_{e \in G_l} \mathcal{A}(e), \; \forall \; G_l \in G
\end{equation}
\vspace{-0.3cm}
\begin{equation}\label{eq:rel_update_merging_only}
\mathcal{T}^{*} = \{ \phi(g(e_1)),r,\phi(g(e_2)) \mid (e_1,r,e_2) \in \mathcal{T}, \phi(g(e_1))\neq \phi(g(e_2))\}.
\end{equation}
If the merged description of a canonical entity becomes too long, we summarize it to prevent overly long inputs from a single entity during retrieval. The merge of similar entities effectively reduces the storage cost. However, because numerous modifications are made to the original entity and relation sets, the quality of the resulting knowledge graph largely depends on the effectiveness of the entity embedding or matching methods used.

(2) Synonym Linking Only.
This approach add a synonym relation $r_{\text{syn}}$ between merged entity $e'$ and canonical entity $\phi(G_l)$ inside each group $G_l$ without the modification of entity set and attributes, which can be described as:
\begin{equation}\label{eq:rel_update_synonym_only}
   \mathcal{T}^* = \mathcal{T} \;\cup\; \{(e', r_{\text{syn}}, \phi(G_l)) \mid e' \in G_l \setminus \phi(G_l),\, G_l \in G_{\text{ent}}\}. 
\end{equation}
This method keeps the minimal changes to the original KG $\mathcal{G}$, yet still cannot well resolve duplication of conceptually similar entities inside $\mathcal{G}$, leading to redundancy and low-efficiency during retrieval.

(3) Merging with Synonym Linking.
To prevent the information loss of merged entities as in directly merging, inside each group $G_l$, this approach merges attributes and relations to the canonical entity $\phi(G_l)$ first, then adds synonym relations $r_{\text{syn}}$ towards canonical entity $\phi(G_l)$. In this case, the entity set $\mathcal{E}$ remains unchanged, the relation set $\mathcal{R}$ is updated by Equation (\ref{eq:rel_update_merging_only}), then Equation (\ref{eq:rel_update_synonym_only}), and the attributes is updated by Equation (\ref{eq:attr_merge}).

\vspace{-0.2cm}
\subsection{Triple Reflection}\label{sec:triple_reflection}
\vspace{-0.2cm}

Since the external information in the documents may contain erroneous content, the triples extracted by LLMs are not always trustworthy \citep{llm_error,llm_error2}. Besides, due to the batched generation of name-entity recognition of chunks, errors may also occur \citep{llm_batch_error}. Therefore, we use LLM-as-judge to remove the low-quality triple. Specifically, given a triple, composed of source entity, relation, and target entity, we let LLM to predict a reliability score $s=\mathcal{M}_{\text{judge}}(e_1,r,e_2)$. Then, we filter out the triples that are below a threshold $\delta_\text{TR}$ and the final relation set that we obtain is
\begin{equation}
\mathcal{T}^* = \{ (e_1, r, e_2) \mid (e_1, r, e_2)\in\mathcal{T}, \mathcal{M}_{\text{judge}}(e_1,r,e_2) \ge \delta_\text{TR}\}
\end{equation}
\vspace{-0.5cm}
\subsection{Analysis}
\vspace{-0.2cm}
Under the construction of KGs in Section~\ref{sec:preliminaries}, if no entity resolution is applied, \ie the deduplication function becomes identity function, yielding a union of subgraphs with no cross edges. Retrieval over such a disconnected graph reduces to selecting the information of independent triples that a vanilla retriever would select. Formally, we summarize the claim in Proposition 1 as below, where the proof is provided in Appendix~\ref{apd:sec_proof}.

\vspace{-0.2cm}
\begin{theorem} 1
Given a graph-based RAG and a vanilla RAG system that share the same augmentation and generation processes, the absence of entity resolution causes the graph-based RAG to degrade into vanilla RAG.
\end{theorem}
\vspace{-0.2cm}

Proposition 1 demonstrates that any benefit of Graph-based RAG over vanilla RAG necessarily comes from the connectivity created by entity resolution. 

%% file: sections/Experiments.tex
\vspace{-0.2cm}
\section{Experiments}
\vspace{-0.2cm}

In this section, we comprehensively evaluate the effectiveness the denoising approach mentioned in the previous section for Graph-based RAG systems. We first introduce the experimental settings in Section~\ref{sec:setup}. Then, we demonstrate that entity resolution can significantly reduce the scale of the original graph while improving question-answering performance on Graph-based RAG systems in Section~\ref{sec:impact_entity_resolution}. In Section~\ref{sec:component_analysis}, we test and analyze how different components in entity resolution influence the overall performance. we study the impact of entity reduction ratio and relation reduction ratio on the performance of Graph-based RAG in Section~\ref{sec:hyperparameter_analysis}. Then, we conduct an ablation study in Section~\ref{sec:ablation_study} to evaluate the impact of different deletion methods and LLM API. 
Additional, we conduct a detailed case study in Appendix~\ref{sec:case_study} to illustrate the qualitative differences between knowledge graphs before and after the denoising process.

\vspace{-0.2cm}
\subsection{Experimental Setup}\label{sec:setup}
\vspace{-0.2cm}

\paragraph{Datasets and metrics} We evaluate the performance of Graph-based RAG on four datasets from UltraDomain benchmark \citep{ultrdomain} following \citep{lightrag}, including \textit{Agriculture, CS, Legal}, and \textit{Mix}. \textit{Agriculture, CS,} and \textit{Legal} contains domain-specific knowledge, while \textit{Mix} includes a broad spectrum of disciplines. Please refer to Appendix \ref{apd:sec_data_statistics} for details of data statistics. Different Graph-based RAG systems are tested by question-answering tasks. We use an LLM as a judge to conduct pairwise comparisons between the responses of two methods, where a winning rate greater than $50\%$ indicates that one method outperforms the other, and vice versa. The evaluation considers four dimensions: comprehensiveness, diversity, empowerment, and overall quality. The detailed evaluation process is shown in Appendix \ref{apd:sec_evaluation}.

\textbf{Baselines} We select four popular Graph-based RAG methods as our baselines: (1) LightRAG \citep{lightrag}. 
(2) HippoRAG \citep{hipporag}. 
(3) LGraphRAG \citep{ms_graphrag}. 
(4) GGraphRAG \citep{ms_graphrag}. 

\textbf{Implementation details.} We implement our experiment based on DIGIMON \citep{DIGIMON}, which is a framework that stably implements many variants of Graph-based RAG and provide a fair and unified comparison among these methods. For efficient indexing and retrieval, the entities and relations are stored in vector dataset bases implemented by Llama Index \citep{llama_index}. We use open sourced Qwen3-235B-A22B-Instruct-2507 \citep{qwen3} for the LLM API calling, which natively supports 256K context. The model is deployed using VLLM \citep{vllm} on a Linux server with 8 H20 GPUs. We use Qwen3-Embedding-8B \citep{qwen3embedding} as the embedding model during index building and semantic blocking. For the KG embedding, we use pykeen \citep{pykeen}, which is design for many types of KG embedding. By default, we set the entity reduction ratio as $40\%$ of the total size of the entity set, $\delta_\text{TR}$ of triple reflection as $0.2$, semantic-based method for blocking, LLM embeddings for entity embedding, ego-based similarity for matching, direct merging in merging step. Please refer to Appendix \ref{apd:sec_experimental_details} for more implementation details.
\input{tables/denoising_KG_main}
\vspace{-0.2cm}
\subsection{Impact of Knowledge Graph Denoising}\label{sec:impact_entity_resolution}
\vspace{-0.2cm}

To validate the effectiveness of our proposed \method, we compare the performance of baseline Graph-based RAG with denoised KGs and original KGs on four datasets. As shown in Table \ref{tab:denoising_KG_main}, after reducing $40\%$ of the entities and removing erroneous relations, the performance of Graph-based RAG on cleaned KGs is better than the original KGs in most cases. This indicates the necessity of denoising KGs for Graph-based RAG. Note that for HippoRAG, the performance is not significantly improved on the \textit{Legal} and \textit{Mix} datasets. This is because the entity set of the KG in HippoRAG only contains entity names without descriptions, limiting the performance of entity resolution. 
\vspace{-0.2cm}
\subsection{Component Analysis of Entity Resolution}\label{sec:component_analysis}
\vspace{-0.2cm}
\begin{figure}[htbp]
    \centering
    \includegraphics[width=0.85\linewidth]{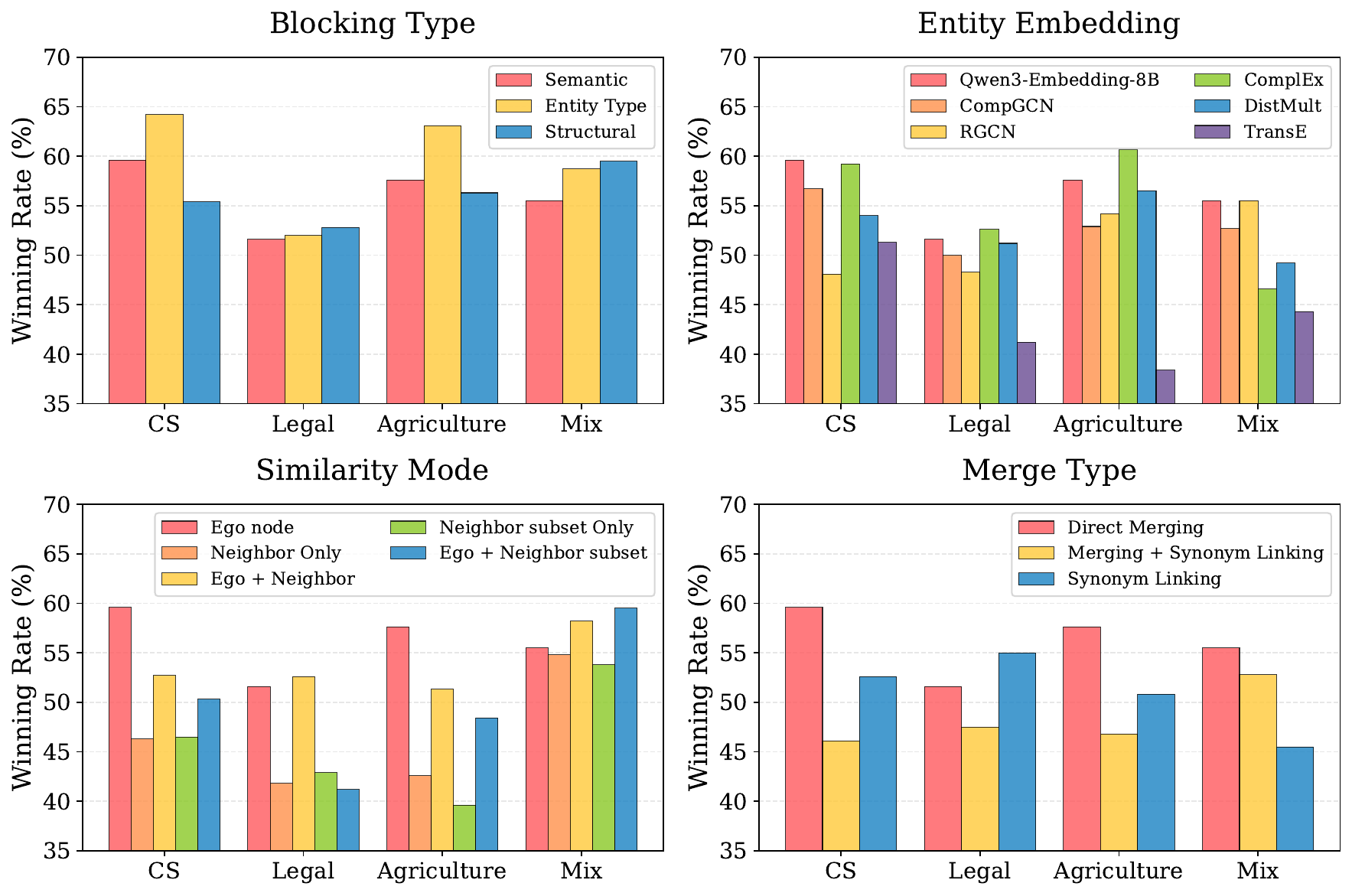}
    \caption{Impact of different entity resolution components on Graph-based RAG performance.}
    \label{fig:component_analysis}
\end{figure}

We further study the impact of different components of entity resolution on the performance of Graph-based RAG. Figure \ref{fig:component_analysis} shows the winning rate averaged across four metrics (Comprehensive, Diversity, Empowerment, and Overall) on denoised KGs with different components of blocking type, entity embedding, similarity mode, and merge type. We find that: (1) Entity type-based blocking is more effective than semantic-based or structure-based blocking. We speculate that entity type is a better and more natural inductive bias for entity resolution and can lead to more robust denoised graph, which is important for graph mining~\citep{luan2022revisiting, zheng2024missing}. (2) Traditional KG embeddings can rival LLM embeddings. In the \textit{Legal} and \textit{Agriculture} datasets, LLM embeddings underperform ComplEx embeddings~\citep{ComplEx}, which represents entities and relations as vectors in a complex number vector space to better handle asymmetric relations. This demonstrates that traditional KG embeddings can be a viable alternative to LLM embeddings, especially in scenarios where computational resources are insufficient for LLMs or when we contain complex relations in the datasets. (3) Without ego-based similarity, the performance of Graph-based RAG degrades in most cases. Additionally, incorporating neighbor information as a complement to ego node information improves performance in the \textit{Legal} and \textit{Mix} datasets. (4) Simple direct merging often surpasses synonym linking. Although both methods aim to deal with the synonym entities, synonym linking only adds synonym relations between merged entities and the canonical entity. As a result, the KGs remain redundant, requiring more hops to retrieve relevant information. In contrast, direct merging addresses this by consolidating entities with similar meanings into a single entity, which is more efficient.

\subsection{Hyperparameter Analysis}\label{sec:hyperparameter_analysis}

\begin{figure}[htbp]
    \centering
    \includegraphics[width=1.0\linewidth]{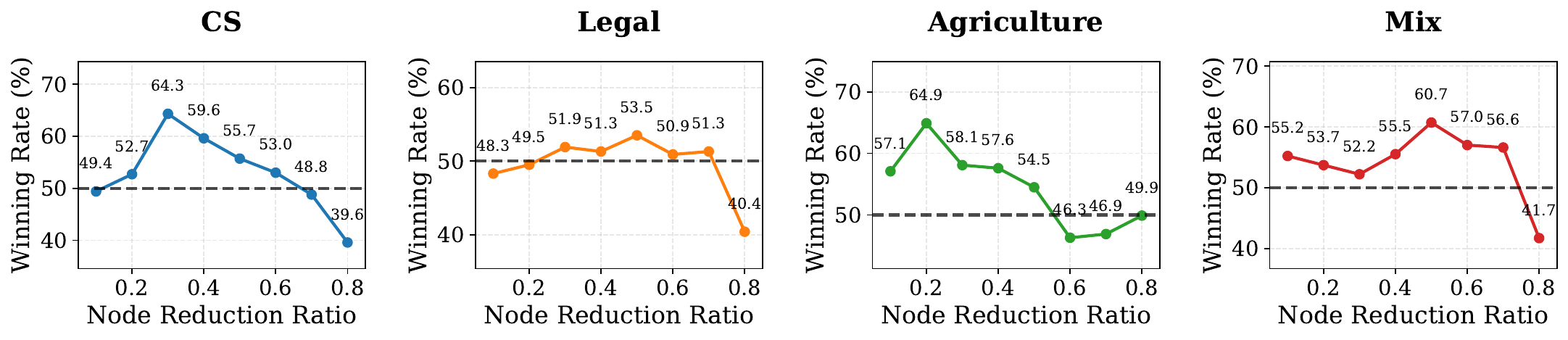}
    \vspace{-0.3cm}
    \caption{Influence of entity reduction ratio on Graph-based RAG performance.}
    \label{fig:node_reduction}
\end{figure}

We conduct experiments to investigate the robustness of the selection of the entity reduction ratio on the effectiveness of denoising. As shown in Figure \ref{fig:node_reduction}, the winning rate is equal or larger than 50\% as long as reduction ratio is not too high. This means, as long as entities are not over-merged, the denoising step is effective for Graph-based RAG.
Notably, on \textit{Mix} and \textit{Legal}, the performance remains comparable to the original KG up to $70\%$, which means even the reduction of  $70\%$ entities in KG does not cause negative effect compared to original KG. 
At such aggressive denoising setting, not only near-duplicate or synonymous entities are merged, but entities with only marginal semantic similarity and overlapping local neighborhoods can also be absorbed into a single canonical node, effectively collapsing fine-grained clusters. The resulting KG becomes substantially more compact while still keep, and sometimes even improve, Graph-based RAG performance. We attribute this to the reduced redundancy, shorter multi-hop paths, and the concentration on fewer, more informative nodes. This indicates that Graph-based RAG is robust to some over-merging cases so long as coarse-grained semantics are preserved.
\vspace{-0.2cm}
\subsection{Ablation Study}\label{sec:ablation_study}
\vspace{-0.2cm}

\begin{wrapfigure}{r}{0.5\textwidth} 
    \centering
    \vspace{-0.5cm} 
    \includegraphics[width=0.48\textwidth]{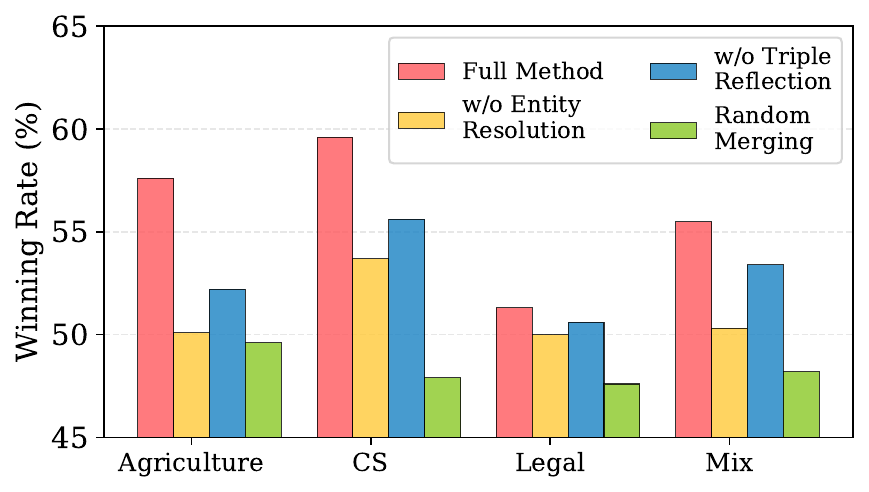}
    \vspace{-0.5cm}
    \caption{Ablation study on the performance of the full denoising method against versions without entity resolution, without triple reflection, and with random entity merging.}
    \vspace{-0.3cm}
    \label{fig:ablation_random_merge}
\end{wrapfigure}
To evaluate the effectiveness of entity resolution and triple reflection in \method, we conduct an ablation study in this subsection. As shown in Figure \ref{fig:ablation_random_merge}, without entity resolution or triple reflection, the performance of Graph-based RAG significantly degrades in all datasets. Moreover, we find that entity resolution is more impactful than triple reflection, indicating the necessity of entity resolution in KGs.  We also set up random merging as a reference method for comparison and the results show worse performance than the above two partial methods, which again shows the necessity to handle the redundant entities smartly.


%% file: tables/denoising_KG_main.tex
\begin{table*}[htbp]
\centering
\label{tab:denoising_KG_main}
\caption{Performance comparison of graph-based RAG methods on original and cleaned knowledge graphs across four datasets. The evaluation is based on winning rates by comparing responses generated from original versus cleaned knowledge graphs.}
\resizebox{\textwidth}{!}{
\begin{tabular}{llcccccccc}
\toprule
\multirow{2}{*}{Dataset} & \multirow{2}{*}{Dimension} 
& \multicolumn{2}{c}{LightRAG} 
& \multicolumn{2}{c}{HippoRAG} 
& \multicolumn{2}{c}{LGraphRAG} 
& \multicolumn{2}{c}{GGraphRAG} \\
\cmidrule(lr){3-4} \cmidrule(lr){5-6} \cmidrule(lr){7-8} \cmidrule(lr){9-10}
 & & Orig. & Clean & Orig. & Clean & Orig. & Clean & Orig. & Clean \\
\midrule
\multirow{4}{*}{Agriculture} 
& Comprehensive & 43.60\% & \textbf{56.40\%} & 49.80\% & \textbf{50.20\%} & 48.80\% & \textbf{51.20\%} & 47.79\% & \textbf{52.21\%} \\
& Diversity     & 41.60\% & \textbf{58.40\%} & 43.78\% & \textbf{56.22\%} & 40.00\% & \textbf{60.00\%} & 36.14\% & \textbf{63.86\%} \\
& Empowerment   & 42.00\% & \textbf{58.00\%} & 47.39\% & \textbf{52.61\%} & 45.60\% & \textbf{54.40\%} & 47.79\% & \textbf{52.21\%} \\
& Overall       & 42.40\% & \textbf{57.60\%} & 48.19\% & \textbf{51.81\%} & 47.20\% & \textbf{52.80\%} & 47.39\% & \textbf{52.61\%} \\
\midrule
\multirow{4}{*}{CS} 
& Comprehensive & 39.20\% & \textbf{60.80\%} & 49.17\% & \textbf{50.83\%} & 47.18\% & \textbf{52.82\%} & 48.19\% & \textbf{51.81\%} \\
& Diversity     & 40.00\% & \textbf{60.00\%} & 35.54\% & \textbf{64.46\%} & 43.55\% & \textbf{56.45\%} & 44.58\% & \textbf{55.42\%} \\
& Empowerment   & 40.80\% & \textbf{59.20\%} & 49.17\% & \textbf{50.83\%} & 47.58\% & \textbf{52.42\%} & 48.59\% & \textbf{51.41\%} \\
& Overall       & 41.60\% & \textbf{58.40\%} & 49.59\% & \textbf{50.41\%} & 46.77\% & \textbf{53.23\%} & 48.19\% & \textbf{51.81\%} \\
\midrule
\multirow{4}{*}{Legal} 
& Comprehensive & 43.60\% & \textbf{50.80\%} & 49.60\% & \textbf{50.40\%} & 44.80\% & \textbf{55.20\%} & 48.00\% & \textbf{52.00\%} \\
& Diversity     & 41.60\% & \textbf{51.20\%} & 44.00\% & \textbf{56.00\%} & 36.80\% & \textbf{63.20\%} & 42.80\% & \textbf{57.20\%} \\
& Empowerment   & 42.00\% & \textbf{51.60\%} & 50.00\% & 50.00\% & 45.20\% & \textbf{54.80\%} & 48.00\% & \textbf{52.00\%} \\
& Overall       & 42.40\% & \textbf{51.60\%} & 50.00\% & 50.00\% & 44.80\% & \textbf{55.20\%} & 47.60\% & \textbf{52.40\%} \\
\midrule
\multirow{4}{*}{Mix} 
& Comprehensive & 45.60\% & \textbf{54.40\%} & 48.80\% & \textbf{51.20\%} & 45.20\% & \textbf{54.80\%} & 49.60\% & \textbf{50.40\%} \\
& Diversity     & 40.80\% & \textbf{59.20\%} & 51.60\% & 48.40\% & 38.40\% & \textbf{61.60\%} & 45.20\% & \textbf{54.80\%} \\
& Empowerment   & 45.60\% & \textbf{54.40\%} & 47.60\% & \textbf{52.40\%} & 42.40\% & \textbf{57.60\%} & 49.20\% & \textbf{50.80\%} \\
& Overall       & 46.00\% & \textbf{54.00\%} & 48.40\% & \textbf{51.60\%} & 42.40\% & \textbf{57.60\%} & 49.40\% & \textbf{50.60\%} \\
\bottomrule
\end{tabular}
}
\end{table*}

%% file: appendix/data_statistics.tex
\section{Data Statistics}\label{apd:sec_data_statistics}

\input{tables/dataset_statistics}

As shown in Table~\ref{tab:dataset_statistics}, we report the numbers of tokens, documents, and questions for the four datasets used in this paper. We also present the counts of entities and relations, as well as the average length of entity descriptions (in tokens) in the LLM-generated knowledge graphs extracted by LightRAG \citep{lightrag}, HippoRAG \citep{hipporag}, LGraphRAG \citep{ms_graphrag}, and GGraphRAG \citep{ms_graphrag}. Note that the knowledge graphs generated by HippoRAG do not contain entity descriptions.

%% file: tables/dataset_statistics.tex
\begin{table*}[ht]
\centering
\label{tab:dataset_statistics}
\caption{Statistics of datasets and knowledge graphs across four domains.}
\resizebox{0.8\textwidth}{!}{
\begin{tabular}{llcccc}
\toprule
\multicolumn{2}{l}{Category} & Agriculture & CS & Legal & Mix \\
\midrule
\multicolumn{2}{l}{\# Token}     & 1,949,526 & 2,047,866 & 4,872,343 & 611,161 \\
\multicolumn{2}{l}{\# Document}  & 12        & 10        & 94        & 61 \\
\multicolumn{2}{l}{\# Question}  & 125       & 125       & 125       & 125 \\
\midrule
\multirow{4}{*}{\# Entity} 
& LightRAG   & 21,131 & 16,434 & 16,502 & 8,942 \\
& HippoRAG   & 42,444 & 25,495 & 34,342 & 24,055 \\
& LGraphRAG  & 21,761 & 15,257 & 16,761 & 10,240 \\
& GGraphRAG  & 21,227 & 15,600 & 16,111 & 10,399 \\
\midrule
\multirow{4}{*}{\# Relation} 
& LightRAG   & 23,102 & 20,642 & 33,625 & 7,458 \\
& HippoRAG   & 41,636 & 25,170 & 51,031 & 16,370 \\
& LGraphRAG  & 25,834 & 19,980 & 36,742 & 8,513 \\
& GGraphRAG  & 21,408 & 19,412 & 36,507 & 9,943 \\
\midrule
\multirow{4}{*}{\shortstack{Ave. Entity \\Description}}
& LightRAG   & 40.47 & 42.12 & 63.64 & 32.61 \\
& HippoRAG   & --    & --    & --    & -- \\
& LGraphRAG  & 40.23 & 40.21 & 62.11 & 31.88 \\
& GGraphRAG  & 38.74 & 39.83 & 63.76 & 33.66 \\
\bottomrule
\end{tabular}
}
\end{table*}

%% file: appendix/more_experiments.tex
\section{Additional Experimental Results}

\subsection{Impact of Different LLMs in RAG}

\input{tables/different_llm_api}

To show how different LLMs backbones influences the performance of \method shown in Table \ref{tab:denoising_KG_main}, apart from Qwen3-235B-A22B \citep{qwen3}, we further conduct experiments using GPT-4o-mini \citep{gpt4omini} and Gemini-2.5-flash \citep{gemini25} on four datasets on LightRAG \citep{lightrag}. As shown in Table \ref{tab:different_llm_api}, under the entity reduction of $40\%$ and triple reflection threshold of $0.2$, the winning rate of using GPT-4o-mini or Gemini-2.5-flash is comparable as Qwen3-235-A22B, indicating the generality of \method across different types of LLMs.

\subsection{Comparison of Token Consumption}

\input{tables/token_cost}

We further compare the costs of \method under different entity reduction ratios. Table \ref{tab:token_statistics} shows the statistics of token consumption after applying \method in LightRAG as shown in Table \ref{tab:denoising_KG_main}. First, we can see that there is no significant differences of  token consumption in prompt and completion for LightRAG on the original knowledge graph and knowledge graphs with \method, indicating the performance gain is not caused by additional information. Second, we notice that the input token increases with node reduction of $20\%$ or $40\%$, then decreases on $60\%$ and $80\%$. We explain this as, in lower reduction ratio, few entities are merged, which slightly increases the input prompt, while in high reduction ratio, more and more entities are merged together, after the summaziation of entitiy description, the total retrieved entites and relations become fewer, leads to fewer input token.

\subsection{Case Study}\label{sec:case_study}
\begin{figure}[h]
    \centering
    \includegraphics[width=0.8\linewidth]{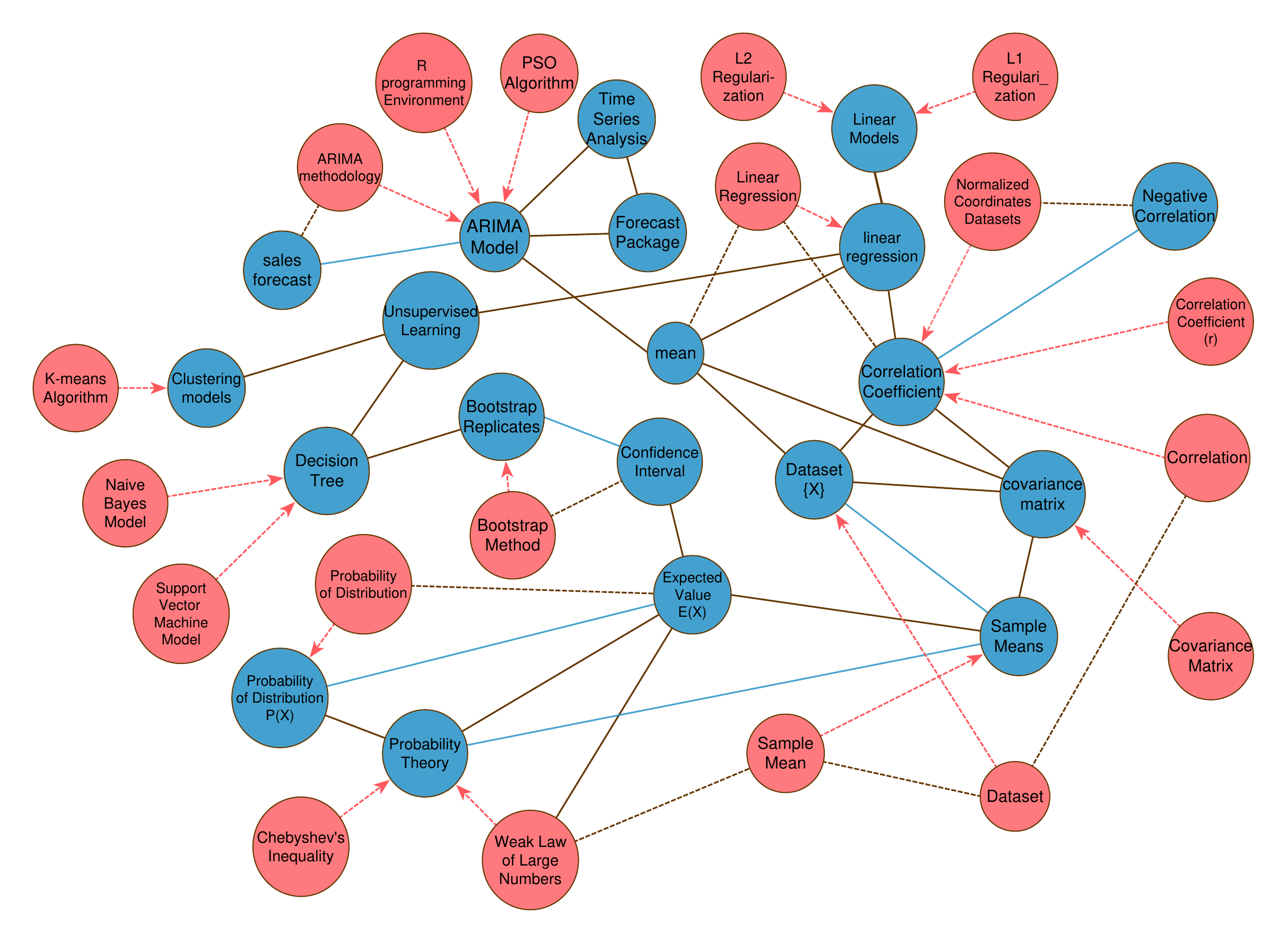}
    \caption{Case Study of Knowledge Graph Denoising on the CS Dataset. The figure illustrates a subgraph before and after applying our denoising method. Redundant entities are denoted in red and merging process is shown in arrows.}
    \label{fig:case_study}
\end{figure}

To illustrate the qualitative impact of denoising, we conduct a case study on entity resolution using the CS dataset. Figure~\ref{fig:case_study} shows a subgraph of the knowledge graph before and after denoising. Red nodes indicate redundant entities that have been merged into their canonical forms, while blue nodes represent entities that remain unchanged. Dashed red lines indicate the direction of merging from one entity to another, green lines denote newly added relations, brown dashed lines represent removed relations, and black lines correspond to relations that are retained.

The entity merging process is generally reasonable. For example, variations such as \texttt{ARIME methodology} are merged into \texttt{ARIMA model}, and \texttt{Linear Regression} into \texttt{linear regression}. We also observe merges driven by semantic similarity, such as \texttt{K-means Algorithm} being merged into \texttt{Clustering models}, and \texttt{Naive Bayes Model} and \texttt{Support Vector Machine Model} being merged into \texttt{Decision Tree}. Overall, the denoised knowledge graph is more concise and efficient, thereby improving the performance of graph-based RAG.

We also examine the cases of triple reflection. As in Table \ref{tab:unreasonable_relations}, we listed some triples with $\delta_\text{TR}\leq0.2$.

\begin{table}[h]
    \centering
    \caption{Case study of triple reflection}
    \label{tab:unreasonable_relations}
    \begin{tabular}{p{2.5cm}|p{2.5cm}|c|p{2.5cm}|p{4.5cm}}
        \hline
        \textbf{Relation} & \textbf{Source} & \textbf{Score} & \textbf{Target} & \textbf{Analysis} \\
        \hline
        Transenterix Inc. owns Safestitch LLC & Transenterix Inc. owns Safestitch LLC, indicating a parent subsidiary relationship & 0.1 & Safestitch LLC & TransEnterix does not own SafeStitch \\
        \hline
        Turtle is one of the entities classified as a borrower & Turtle is one of the entities classified as a borrower in the financial agreement & 0.1 & Borrowers & Turtles are not entities that engage in borrowing \\
        \hline
        Michael Scott is involved in the SEC lawsuit & Michael Scott is involved in the SEC lawsuit as a defendant accused of securities violations & 0.1 & SEC lawsuit & Michael Scott is a fictional character from the television show 'The Office' and not a real person involved in any legal matters \\
        \hline
        Title policy for Pabst & Title policy is required to obtain a title policy to ensure the legitimacy of the asset ownership during the acquisition & 0.1 & Pabst & A title policy is a type of insurance related to real estate transactions, while 'pabst' appears to refer to a brand \\
        \hline
        Shareholder's equity reflects net worth of dealers & Shareholder's equity is a key financial metric that reflects the net worth of dealers after liabilities are deducted & 0.2 & Dealers & Shareholder's equity is a financial metric relevant to companies and their owners, not specifically to dealers \\
        \hline
        Kristen M Jenner and Kylie K Jenner are key executives & Kylie K Jenner and Kristen M Jenner are both identified as key executives, indicating a professional relationship in a business context & 0.2 & Kylie K Jenner & Kristen M Jenner is not a recognized executive in the same context as Kylie K Jenner \\
        \hline
    \end{tabular}
\end{table}

%% file: tables/different_llm_api.tex
\begin{table*}[ht]
\centering
\label{tab:different_llm_api}
\caption{Performance comparison of models on original and cleaned knowledge graphs across four datasets. The evaluation is based on winning rates by comparing responses generated from original versus cleaned knowledge graphs.}
\resizebox{0.92\textwidth}{!}{
\begin{tabular}{llcccccc}
\toprule
\multirow{2}{*}{Dataset} & \multirow{2}{*}{Dimension} 
& \multicolumn{2}{c}{Qwen3-235B-A22B} 
& \multicolumn{2}{c}{GPT-4o-mini} 
& \multicolumn{2}{c}{Gemini-2.5-flash} \\
\cmidrule(lr){3-4} \cmidrule(lr){5-6} \cmidrule(lr){7-8}
 & & Orig. & Clean & Orig. & Clean & Orig. & Clean \\
\midrule
\multirow{4}{*}{Agriculture} 
& Comprehensive & 43.60\% & \textbf{56.40\%} & 45.34\% & \textbf{54.66\%} & 46.00\% & \textbf{54.00\%} \\
& Diversity     & 41.60\% & \textbf{58.40\%} & 29.27\% & \textbf{70.73\%} & 46.00\% & \textbf{54.00\%} \\
& Empowerment   & 42.00\% & \textbf{58.00\%} & 31.71\% & \textbf{68.29\%} & 46.80\% & \textbf{53.20\%} \\
& Overall       & 42.40\% & \textbf{57.60\%} & 33.74\% & \textbf{66.26\%} & 46.80\% & \textbf{53.20\%} \\
\midrule
\multirow{4}{*}{CS} 
& Comprehensive & 39.20\% & \textbf{60.80\%} & 42.32\% & \textbf{57.68\%} & 44.40\% & \textbf{55.60\%} \\
& Diversity     & 40.00\% & \textbf{60.00\%} & 36.51\% & \textbf{63.49\%} & 43.20\% & \textbf{55.60\%} \\
& Empowerment   & 40.80\% & \textbf{59.20\%} & 41.91\% & \textbf{58.09\%} & 43.20\% & \textbf{56.80\%} \\
& Overall       & 41.60\% & \textbf{58.40\%} & 41.91\% & \textbf{58.09\%} & 44.00\% & \textbf{56.00\%} \\
\midrule
\multirow{4}{*}{Legal} 
& Comprehensive & 43.60\% & \textbf{56.40\%} & 46.40\% & \textbf{53.60\%} & 42.00\% & \textbf{58.00\%} \\
& Diversity     & 41.60\% & \textbf{58.40\%} & 45.20\% & \textbf{54.80\%} & 42.40\% & \textbf{57.60\%} \\
& Empowerment   & 42.00\% & \textbf{58.00\%} & 46.80\% & \textbf{53.20\%} & 40.80\% & \textbf{59.20\%} \\
& Overall       & 42.40\% & \textbf{57.60\%} & 47.60\% & \textbf{52.40\%} & 41.20\% & \textbf{58.80\%} \\
\midrule
\multirow{4}{*}{Mix} 
& Comprehensive & 45.60\% & \textbf{54.40\%} & 47.18\% & \textbf{52.82\%} & 42.40\% & \textbf{57.60\%} \\
& Diversity     & 40.80\% & \textbf{59.20\%} & 43.95\% & \textbf{56.05\%} & 40.00\% & \textbf{60.00\%} \\
& Empowerment   & 45.60\% & \textbf{54.40\%} & 45.16\% & \textbf{54.84\%} & 42.40\% & \textbf{57.60\%} \\
& Overall       & 46.00\% & \textbf{54.00\%} & 45.56\% & \textbf{54.44\%} & 42.00\% & \textbf{58.00\%} \\
\bottomrule
\end{tabular}
}
\end{table*}

%% file: tables/token_cost.tex
\begin{table*}[ht]
\centering
\label{tab:token_statistics}
\caption{Token consumption statistics under different entity reduction ratios across four datasets.}
\resizebox{0.92\textwidth}{!}{
\begin{tabular}{llccccc}
\toprule
Dataset & Type & Original & 20\% & 40\% & 60\% & 80\% \\
\midrule
\multirow{3}{*}{Mix} 
& Prompt       & 1,040,189 & 1,185,787 & 1,267,955 & 1,149,659 & 1,133,338 \\
& Completion   &   86,171  &   85,738  &   85,454  &   85,334  &   86,051  \\
& Total Token  & 1,126,360 & 1,271,525 & 1,353,409 & 1,234,993 & 1,219,389 \\
\midrule
\multirow{3}{*}{CS} 
& Prompt       & 1,084,623 & 1,118,326 & 1,106,513 &   906,618 &   779,191 \\
& Completion   &   89,056  &   90,658  &   89,394  &   88,844  &   89,252  \\
& Total Token  & 1,173,679 & 1,208,984 & 1,195,907 &   995,462 &   868,443 \\
\midrule
\multirow{3}{*}{Agriculture} 
& Prompt       & 1,273,710 & 1,537,191 & 1,296,947 & 1,278,717 &   911,124 \\
& Completion   &   82,351  &   82,677  &   82,978  &   79,724  &   79,683  \\
& Total Token  & 1,356,061 & 1,619,868 & 1,379,925 & 1,358,441 &   990,807 \\
\midrule
\multirow{3}{*}{Legal} 
& Prompt       & 1,755,056 & 1,749,183 & 1,721,740 & 1,528,838 & 1,658,700 \\
& Completion   &   84,124  &   84,771  &   84,178  &   83,707  &   85,468  \\
& Total Token  & 1,839,180 & 1,833,954 & 1,805,918 & 1,612,545 & 1,744,168 \\
\bottomrule
\end{tabular}
}
\end{table*}

%% file: appendix/experimental_details.tex
\section{Implementation Details}\label{apd:sec_experimental_details}

\subsection{Graph-based RAG}

For all the Graph-based RAG methods, we set token-based chunking across all methods, with segment length of approximately 1,200 tokens and an overlap of 100 tokens, using a standard tokenizer to balance context preservation and indexing granularity. We set the retriever to return the top 5 candidates. When personalized PageRank is used, we set entity-aware priors with light damping to encourage focus on salient nodes. All methods answer questions directly rather than only returning supporting context. We set the overall candidate pool to 20. We set token budgets consistently across methods: the naive assembly budget to 12,000 tokens, the local assembly budget to 4,000 tokens, and the entity and relation evidence budgets to 2,000 tokens each. When iterative reasoning over retrieved evidence is enabled, we cap the refinement steps at 2.

LightRAG \citep{lightrag} maintains both entity and relation indices and builds a relation-centric knowledge graph enriched with edge keywords. We enable entity descriptions, entity types, edge descriptions, and edge names to maximize semantic coverage. We set the usable context window to 32,768 tokens. For retrieval, we set nearest-neighbor search and enable entity-similarity–aware propagation with the top 5 results. Querying is hybrid: we enable both local and global graph search. We set the global community cap to 512 without a minimum rating, the global community report budget to 16,384 tokens, and the global context budget to 4,000 tokens. Locally, we set the context budget to 4,800 tokens and the community report budget to 3,200 tokens. We allow keyword cues when composing the final context.

HippoRAG \citep{hipporag} focuses on an entity–relation graph with entity-link–aware chunking and enables graph augmentation while keeping metadata conservative: we disable entity and edge descriptions, and we retain edge names. We set retrieval to personalized PageRank over the entity–relation graph without an entity-similarity term in propagation, and we set the top-$k$ to 5. Querying follows a hybrid strategy while we disable explicit propagation-based augmentation in the final context assembly. We keep the same token budgets as in the common configuration, and we cap iterative reasoning at 2 steps.

LGraphRAG \citep{ms_graphrag} uses a relation-centric knowledge graph with a forced construction setting. We enable entity and edge descriptions and edge names, and we disable entity types. We apply community-aware clustering using the Leiden algorithm; we set the maximum community size to 10 and use concise community summaries. We set retrieval to nearest-neighbor search with an additional local neighborhood expansion, and we enable propagation-based augmentation while disabling global community selection. We set the local context budget to 4,800 tokens and the local community report budget to 3,200 tokens, and we keep the same overall budgets and refinement limits as in the common setup.

GGraphRAG \citep{ms_graphrag} adopts the same relation-centric graph construction and community-aware clustering as LGraphRAG. We set retrieval to nearest-neighbor search without local expansion, and we enable both local and global querying. We set the global community cap to 512, the global community report budget to 16,384 tokens, and the global context budget to 4,000 tokens, while keeping the local budgets aligned with the common configuration. Other token allocations and refinement limits follow the common setup.

\subsection{Reduction Ratio}

We further report the number and proportion of removed entities and relations in Table \ref{tab:denoising_KG_main}. As shown in Table \ref{tab:reduction_ratio_number}, across the four datasets, the entity reduction ratio is approximately $40\%$. The relation reduction ratio ranges from $30\%$ to $60\%$, reflecting both the removal of relations during triple reflection and the disappearance of relations associated with merged entities.

\input{tables/actual_reduction_number}

\subsection{Prompts in Entity Resolution}

To avoid the exceeding length of descriptions of merged knoweldge graphs, we summarize the descriptions if the number of token exceed 4,000. We provide the summarization prompt of entity and relation as follows

\begin{tcolorbox}[colback=white,colframe=black,title=Entity description summarization prompt]
You are a helpful assistant. Please summarize the following list of descriptions for the entity \texttt{\{entity\_name\}} into a single, coherent paragraph. Combine the key information and remove redundant details.

Descriptions to summarize: \\
\texttt{\{description\_list\}}

\textbf{Concise Summary:}
\end{tcolorbox}

\begin{tcolorbox}[colback=white,colframe=black,title=Relation description summarization prompt]
You are a helpful assistant. Please summarize the following list of descriptions for the relationship \texttt{\{item\_name\}} into a single, coherent paragraph. Combine the key information and remove redundant details.

Descriptions to summarize: \\
\texttt{\{description\_list\}}

\textbf{Concise Summary:}
\end{tcolorbox}

\subsection{Prompts in Triple Reflection}

We perform triple reflection on knowledge graph triples (edges) using LLMs to assess their reasonableness before downstream use. For each triple, an LLM returns a numerical quality score and a short analysis; results are written as JSONL for subsequent aggregation and filtering.

\medskip
\begin{tcolorbox}[colback=gray!6,colframe=black,title=System prompt]
You are a knowledge graph expert who evaluates whether the knowledge graph triplet belongs to commonsense knowledge.
\end{tcolorbox}

\begin{tcolorbox}[colback=white,colframe=black,title=User prompt]
Evaluate the reasonableness of the knowledge graph triplet with precision: \\[-0.5em]
\begin{flushleft}\small
Source: \texttt{<source>}\\
Destination: \texttt{<destination>}\\
Relationship: \texttt{<relationship>}
\end{flushleft}

\textbf{Analysis requirements}
\begin{itemize}[leftmargin=*]
  \item \textbf{Semantic accuracy}: Does the relationship accurately describe the connection? Consider domain knowledge and factual correctness.
  \item \textbf{Relevance}: Is the connection meaningful and significant, not trivial or coincidental?
  \item \textbf{Specificity}: Is the relationship clear and specific rather than vague or overly general?
  \item \textbf{Logical coherence}: Does the triple follow expected semantic and syntactic patterns for KGs?
  \item \textbf{Entity type compatibility}: Is the relationship sensible given the entity types involved?
\end{itemize}

\textbf{Scoring guidelines}
\begin{itemize}[leftmargin=*]
  \item \(0.0\text{--}0.3\): Invalid or highly questionable (factually wrong, illogical, meaningless)
  \item \(0.4\text{--}0.6\): Partially valid but problematic (some relevance yet vague/imprecise/minor inaccuracies)
  \item \(0.7\text{--}0.8\): Mostly valid (accurate but could be more specific or informative)
  \item \(0.9\text{--}1.0\): Fully valid (accurate, specific, informative, and logically sound)
\end{itemize}

\textbf{Optimization notes}
\begin{itemize}[leftmargin=*]
  \item Focus on direct evaluation without unnecessary elaboration.
  \item Use domain-specific reasoning where applicable.
\end{itemize}

Output format (return a valid JSON object):
\begin{verbatim}
{
    "analysis": "concise analysis",
    "score": 0.5
}
\end{verbatim}
The score should be a float between 0.0--1.0 with two-decimal precision.
\end{tcolorbox}

\subsection{Evaluation}\label{apd:sec_evaluation}
We assess the responses of \method using an LLM judge in a pairwise-comparison setup. For each question the judge receives the question and two candidate answers from original knowledge graphs or denoised knowledge graphs by \method, and decides which answer is better and why. To mitigate position bias we run two passes per question. Pass A uses (Answer~1, Answer~2) and Pass B swaps the order. Aggregated wins for a method on a criterion are computed by summing Answer~1 wins in Pass A and Answer~2 wins in Pass B. Ties are recorded when the judge issues a tie token. The judge receives the following prompts verbatim.

\medskip
\begin{tcolorbox}[colback=gray!6,colframe=black,title=System prompt]
You are an expert tasked with evaluating two answers to the same question based on three criteria: \textbf{Comprehensiveness}, \textbf{Diversity}, and \textbf{Empowerment}.
\end{tcolorbox}

\begin{tcolorbox}[colback=white,colframe=black,title=User prompt]
You will evaluate two answers to the same question using the three criteria below:
\begin{itemize}[leftmargin=*]
  \item \textbf{Comprehensiveness}: How much detail does the answer provide to cover all aspects and details of the question?
  \item \textbf{Diversity}: How varied and rich is the answer in presenting different perspectives and insights?
  \item \textbf{Empowerment}: How well does the answer help the reader understand the topic and make informed judgments?
\end{itemize}
For each criterion, choose the better answer (\textbf{Answer~1} or \textbf{Answer~2}) and explain why. Then select an overall winner based on these three categories.

Here is the question: \texttt{\{query\}}

Here are the two answers: \\
\textbf{Answer 1:} \texttt{\{answer1\}} \\
\textbf{Answer 2:} \texttt{\{answer2\}}

Evaluate both answers using the three criteria above and provide detailed explanations for each criterion.

Output your evaluation in the following JSON format:

\begin{verbatim}
{
    "Comprehensiveness": {
        "Winner": "[Answer 1 or Answer 2]",
        "Explanation": "[Provide explanation here]"
    },
    "Diversity": {
        "Winner": "[Answer 1 or Answer 2]",
        "Explanation": "[Provide explanation here]"
    },
    "Empowerment": {
        "Winner": "[Answer 1 or Answer 2]",
        "Explanation": "[Provide explanation here]"
    },
    "Overall Winner": {
        "Winner": "[Answer 1 or Answer 2]",
        "Explanation": "[Summarize why this answer is the 
        overall winner based on the three criteria]"
    }
}
\end{verbatim}

\end{tcolorbox}

%% file: tables/actual_reduction_number.tex
\begin{table*}[ht]
\centering
\caption{Statistics of original and cleaned knowledge graphs across four datasets and four Graph-based RAG models.}\vspace{0.2cm}
\label{tab:reduction_ratio_number}
\resizebox{1.0\textwidth}{!}{
\begin{tabular}{llcccccccccccc}
\toprule
\multirow{2}{*}{Dataset} & \multirow{2}{*}{Dimension} 
& \multicolumn{3}{c}{LightRAG} 
& \multicolumn{3}{c}{HippoRAG} 
& \multicolumn{3}{c}{LGraphRAG} 
& \multicolumn{3}{c}{GGraphRAG} \\
\cmidrule(lr){3-5} \cmidrule(lr){6-8} \cmidrule(lr){9-11} \cmidrule(lr){12-14}
 & & Orig. & Clean & Reduction & Orig. & Clean & Reduction & Orig. & Clean & Reduction & Orig. & Clean & Reduction \\
\midrule
\multirow{2}{*}{Agriculture} 
& \# Entity   & 21131 & 12679 & 40.00\% & 42444 & 25466 & 40.00\% & 21761 & 13057 & 40.00\% & 21227 & 12736 & 40.00\% \\
& \# Relation & 23102 & 15548 & 32.70\% & 41636 & 20321 & 51.19\% & 25834 & 16503 & 36.12\% & 21408 & 11258 & 47.41\% \\
\midrule
\multirow{2}{*}{CS} 
& \# Entity   & 16434 & 9861 & 40.00\% & 25495 & 15297 & 40.00\% & 15257 & 9154 & 40.00\% & 15600 & 9360 & 40.00\% \\
& \# Relation & 20642 & 12164 & 41.07\% & 25170 & 13801 & 45.17\% & 19980 & 13756 & 33.15\% & 19412 & 13742 & 29.21\% \\
\midrule
\multirow{2}{*}{Legal} 
& \# Entity   & 16502 & 9902 & 40.00\% & 34342 & 20606 & 40.00\% & 16761 & 10057 & 40.00\% & 16111 & 9667 & 40.00\% \\
& \# Relation & 33625 & 21261 & 36.77\% & 51031 & 35920 & 29.61\% & 36742 & 22987 & 37.44\% & 36507 & 14025 & 61.58\% \\
\midrule
\multirow{2}{*}{Mix} 
& \# Entity   & 8942 & 5366 & 40.00\% & 24055 & 14433 & 40.00\% & 10240 & 6144 & 40.00\% & 10399 & 6240 & 40.00\% \\
& \# Relation & 7458 & 5164 & 30.76\% & 16370 & 6896 & 57.87\% & 8513 & 6288 & 26.14\% & 9943 & 6713 & 32.49\% \\
\bottomrule
\end{tabular}
}
\end{table*}

%% file: appendix/proof.tex
\section{Proof of Proposition 1}\label{apd:sec_proof}

\begin{theorem} 1
Given a graph-based RAG and a vanilla RAG system that share the same augmentation and generation processes, the absence of entity resolution causes the graph-based RAG to degrade into vanilla RAG.
\end{theorem}

\textit{Proof.} We assume that: (1) both systems use identical augmentation and generation processes except for the knowledge representation, (2) vanilla RAG retrieves chunks based on relevance scoring, and (3) graph-based RAG retrieves subgraphs or triples based on query-entity matching. This is not a formal proof but rather an intuitive argument.

Given document chunks $\mathcal{C} = \{c_1,\dots,c_M\}$, a Graph-based RAG system constructs a knowledge graph $\mathcal{G}^* = (\mathcal{E}^*,\mathcal{R}^*,\mathcal{T}^*,\mathcal{A}^*)$ through named entity recognition followed by deduplication. The response $\mathcal{Y}$ is generated for query $Q$ as:
\begin{equation}
\mathcal{Y} \;=\; \mathcal{M} \circ \text{Aug}\big[ Q, \text{Ret}(Q,\mathcal{G}^*) \big].
\end{equation}

Without entity resolution, the deduplication function becomes the identity mapping $\phi(e) = e$ for all $e \in \mathcal{E}_{\text{raw}}$. This means:
\begin{align}
\mathcal{E}^* &= \{\phi(e) \mid e \in \mathcal{E}_{\text{raw}}\} = \mathcal{E}_{\text{raw}} \\
\mathcal{T}^* &= \mathcal{T}_{\text{raw}} \\
\mathcal{A}^*(e) &= \mathcal{A}_{\text{raw}}(e) \quad \forall e \in \mathcal{E}^*
\end{align}

Since each triple $(e_1, r, e_2) \in \mathcal{T}_{\text{raw}}$ originates from a single chunk $c_m$, and no entity merging occurs, entities from different chunks remain disconnected even if they represent the same real-world concept. Formally, let $\mathcal{E}_m = \{e_1, e_2 \mid (e_1, r, e_2) \in \mathcal{T}_m\}$ be entities extracted from chunk $c_m$. Without entity resolution, there are no edges connecting entities from different chunks:
\begin{equation}
\forall i \neq j: \quad \mathcal{N}(e_i) \cap \mathcal{E}_j = \emptyset \quad \text{where } e_i \in \mathcal{E}_i
\end{equation}

This results in $M$ disconnected subgraphs $\mathcal{G}_1^*, \mathcal{G}_2^*, \ldots, \mathcal{G}_M^*$, where each $\mathcal{G}_m^* = (\mathcal{E}_m, \mathcal{R}_m, \mathcal{T}_m, \mathcal{A}_m)$ corresponds to chunk $c_m$.

For any query $Q$, the graph retrieval function $\text{Ret}(Q, \mathcal{G}^*)$ can only retrieve from individual disconnected components. Since each component $\mathcal{G}_m^*$ contains only local information from chunk $c_m$, the retrieved content consists of triples $\mathcal{T}_m$ that represent structured partitions of the original chunk content. The graph-based retrieval without entity resolution becomes:
\begin{equation}
\text{Ret}(Q, \mathcal{G}^*) = \bigcup_{m: \text{rel}(Q, \mathcal{G}_m^*) > \tau} \mathcal{T}_m
\end{equation}
where $\text{rel}(Q, \mathcal{G}_m^*)$ measures relevance between query and local subgraph, and $\tau$ is a threshold.

Note that each original chunk $c_m$ can be decomposed as:
\begin{equation}
c_m = \mathcal{T}_m \cup \text{unextracted text}
\end{equation}
where $\mathcal{T}_m$ represents the structured information extracted from $c_m$. Since $\mathcal{T}_m \subset c_m$, the retrieved triples are essentially parts of the original chunks. With no cross-chunk connections, this retrieval process can be considered as a vanilla RAG system:
\begin{equation}
\text{Ret}_{\text{vanilla}}(Q, \{\mathcal{T}_m\}) = \{\mathcal{T}_m \mid \text{rel}(Q, \mathcal{T}_m) > \tau'\}
\end{equation}
for appropriately chosen thresholds $\tau$ and $\tau'$.

Since the augmentation and generation processes are identical by assumption, and the retrieved content has the same information coverage (parts of chunks vs. disconnected subgraphs), we have:
\begin{equation}
\mathcal{Y}_{\text{graph}} = \mathcal{M} \circ \text{Aug}[Q, \text{Ret}(Q,\mathcal{G}^*)] \equiv \mathcal{M} \circ \text{Aug}[Q, \text{Ret}_{\text{vanilla}}(Q,\{\mathcal{T}_m\})] = \mathcal{Y}_{\text{vanilla}}
\end{equation}

Therefore, without entity resolution, graph-based RAG degrades to vanilla RAG.